%% file: main.tex
\documentclass[runningheads]{llncs}
\usepackage[utf8]{inputenc}
\usepackage{times}
\usepackage{multirow}
\usepackage{float}
\usepackage{placeins}
\usepackage[dvipsnames]{xcolor}
\usepackage{csquotes}
\usepackage{cite}
\usepackage{graphicx}
\usepackage{fancyvrb}
\usepackage{soul}
\usepackage{ragged2e}
\usepackage{tabularx}

\usepackage{tikz}
\usepackage{graphicx}
\graphicspath{ {./images/} }
\usepackage{caption}
\usepackage[caption=false]{subfig}
\usepackage{adjustbox}
\usepackage{filecontents}
\usepackage{pgfplots, pgfplotstable}
\pgfplotsset{compat=1.15}
\usepackage{makecell}
\usepackage[justification=centering]{caption}
\tikzstyle{Action}=[rectangle, draw=black, rounded corners, minimum height = 0.8cm, text width=3cm, text centered]
\tikzstyle{Line}=[-stealth, line width=0.5mm]
\usetikzlibrary{decorations.pathreplacing}
\usetikzlibrary{shapes,snakes,shadows,calc,automata,positioning}
\usepackage{listings}
\usepackage{color}
\definecolor{dkgreen}{rgb}{0,0.6,0}
\definecolor{gray}{rgb}{0.5,0.5,0.5}
\definecolor{mauve}{rgb}{0.58,0,0.82}

\def\SB#1{\textsubscript{#1}}

\newcommand{\errorband}[5][]{ 
\pgfplotstableread{#2}\datatable
    \addplot [draw=none, stack plots=y, forget plot] table [
        x={#3},
        y expr=\thisrow{#4}-\thisrow{#5}
    ] {\datatable};

    \addplot [draw=none, fill=gray!40, stack plots=y, area legend, #1,] table [
        x={#3},
        y expr=2*\thisrow{#5}
    ] {\datatable} \closedcycle;

    \addplot [forget plot, stack plots=y,draw=none] table [x={#3}, y expr=-(\thisrow{#4}+\thisrow{#5})] {\datatable};
}

\begin{document}
\title{Text Data Augmentation: Towards better detection of spear-phishing emails}
\titlerunning{ }

\author{
	Mehdi Regina \and
	 Maxime Meyer\and
	  S\'{e}bastien Goutal
  }

\authorrunning{ }

\institute{Vade Secure\\
\email{\{firstname.lastname\}@vadesecure.com}}

\maketitle

\input{abstract}
\input{introduction}
\input{state-of-the-art}
\input{method} 
\input{experiments} 
\input{arg}
\input{conclusion}
\input{acknowledgments}

\bibliographystyle{splncs04}
\bibliography{bibfile}

\newpage
\appendix
\input{appendix}

\end{document}

%% file: abstract.tex
\begin{abstract}
Text data augmentation, i.e., the creation of new textual data from an existing text, is challenging. 
Indeed, augmentation transformations should take into account language complexity while being relevant to the target Natural Language Processing (NLP) task (e.g., Machine Translation, Text Classification). 
Initially motivated by an application of Business Email Compromise (\textit{BEC}) detection, we propose a corpus and task agnostic augmentation framework used as a service to augment English texts within our company.
Our proposal combines different methods, utilizing BERT language model \cite{devlin2018bert}, multi-step back-translation and heuristics\footnote{Approach to solve a problem employing a practical method that is not guaranteed to be optimal.}.
We show that our augmentation framework improves performances on several text classification tasks using publicly available models and corpora as well as on a \textit{BEC} detection task.
We also provide a comprehensive argumentation about the limitations of our augmentation framework.
\end{abstract}

%% file: introduction.tex
\section{Introduction} 
\label{sec:intro}

Generalization, i.e., the ability of models to predict accurately from unseen data\footnote{Training data as well as unseen data are supposed to be sampled from an unknown distribution.}, is paramount for supervised learning applications.
However, models generalization is often hindered by a lack of data, the acquisition of annotated data being costly.
Data augmentation, which consists in creating new samples from existing ones, helps solve this problem by increasing samples diversity and reinforcing data characteristics such as invariances.

Ubiquitous in computer vision and audio analysis, data augmentation contribution to benchmark results is significant \cite{salamon2017deep, perez2017effectiveness}. 
Data augmentation methods usually involve relatively simple operations (e.g., image rotation, frequency modulation) leveraging data continuity and knowledge about invariances regarding the final task (e.g., object detection, voice recognition).
In NLP however, data augmentation has had limited application.
Guaranteeing augmented texts intelligibility, grammatical correctness and relevance to a given NLP task is a difficult challenge often involving domain or corpus dependent methods \cite{wu2019conditional, kobayashi2018contextual}.  
Indeed, text processing operates in a discrete space of words and copes with natural language complexity (e.g., polysemy, ambiguity, common-sense knowledge \cite{mccarthy1986applications}).

Recently, language representation models \cite{devlin2018bert, yang2019xlnet, lewis2019bart} leveraging transfer learning have emerged, improving model generalization on NLP tasks.
For instance, BERT \cite{devlin2018bert} or XLNet \cite{yang2019xlnet} have been successfully finetuned improving state-of-the-art NLP results without requiring data augmentation.
However, text augmentation still seems relevant in domains where annotated data cost is particularly high.
It is the case of Business Email Compromise (BEC) detection.

BEC are among the most advanced email attacks relying on social engineering and human psychology to trick targeted victims \cite{jakobsson2016understanding, mansfield2016imitation, proofpoint180315}. 
They convey attacks through their textual content (e.g., impersonation, request of payment or sensitive information) almost indistinguishable from legitimate ones, evading traditional email filtering methods \cite{oliver2008fraudulent, levow2009multilevel, idika2007survey, nissim2015detection}.
This elaborate and targeted content drives the rarity of BEC \cite{cidon2019high, ho2017detecting}.
For these reasons, such emails are particularly difficult to detect, collect and can lead to substantial financial losses \cite{fbiICR2020, kwak2020users}.
We address BEC detection with different components to extract and analyze information from an email. 
One of these components includes an email text classifier trained on an annotated corpus (see Section~\ref{sec:experiments}).
This corpus made of BEC and benign email texts is unbalanced because of a limited amount of BEC.

We developed an augmentation framework that can be easily used in different contexts such as our BEC detection use case but also sentiment analysis with Stanford Sentiment Treebank (SST-2) \cite{socher2013recursive} or question classification with Trec REtrieval Conference (TREC-6) \cite{li2002learning}. Examples of augmentations are presented in Table~\ref{table:aug_example}.

\begin{table*}
\centering
\begin{tabularx}{\textwidth}{c||X|X}
\hline
\textbf{Dataset} & Original text   & Augmented text \\
\hline
SST-2 & the entire movie establishes a wonderfully creepy mood  & the whole movie creates a wonderfully scary mood\\
\hline
TREC-6 & What is the secret of the universe?  & What is the mystery of the cosmos?\\
\hline
BEC &  Have you got a minute? I need you to complete a task for me discreetly. P.S : I am in a meeting and can't talk. & Do you have a minute? You have to do a task for me discretly. P.S: I'm IAM and I can't speak.\\
\hline
\end{tabularx}
\caption{Augmentation examples}
\label{table:aug_example}
\end{table*}

In this paper, we propose a corpus and task agnostic framework that can be easily used to augment any English text corpus without the need to train or finetune models.
Our framework leverages an ensemble of methods (Section~\ref{sec:method}) combined for the first time, to the best of our knowledge. 
We assess the contribution of our framework to different datasets in Section~\ref{sec:experiments}.
Finally, we provide a discussion on the limitations of our framework in Section~\ref{sec:arg}.

%% file: state-of-the-art.tex
\section{Related Work}
\label{sec:sota}


Text augmentation is usually applied in a discrete space of words, in which a simple change implies to change a whole word.
Such changes lead to the use of word replacement methods to augment texts.
Those replacements are valid if they preserve data characteristics which can be achieved in supervised learning settings guaranteeing augmented texts compatibility with their labels\footnote{Tag associated to each data sample in regards to a supervised task.}.

Preserving text meaning through the augmentation may be sufficient to satisfy label compatibility in many contexts.
Thus, some works rely on WordNet to find synonyms replacements \cite{giridhara2019study, wei2019eda}, while others leverage word embeddings and latest language models such as BERT \cite{wu2019conditional, kumar2020data}.
Our work benefits from both methods.
Besides word replacements, other operations at word level have been explored such as word dropout or word swap \cite{wei2019eda}. 
However, these operations do not guarantee sentence readability or syntax validity, potentially bringing discrepancy between augmented texts and original ones.

Natural language generation \cite{radford2019language, kumar2020data} can also be used to augment text, however guaranteeing label compatibility implies control over the generation process.
Paraphrase generation has been widely studied and may provide sufficient control over the generation process.
For instance, Gupta et al. \cite{gupta2018deep} use a Variational Autoencoder (VAE) \cite{kingma2013autoencoding, rezende2014stochastic} conditioned on the original text, and Li et al. \cite{li2017paraphrase} propose to use a generator and evaluator network with a supervised training and Reinforcement Learning finetuning.
These works have shown promising results but there is still a gap in terms of readability and relevance between generated and ground truth paraphrases.
Moreover, these architectures can be complex to implement (e.g., several training phases, multiple networks).

An alternative to paraphrase generation is back-translation \cite{edunov2018understanding}.
Machine translation has significantly improved in the last years with the growing use of Neural Machine Translation (NMT) \cite{wu2016google, bojar2016findings}. 
NMT relies on Seq2Seq models \cite{chen2018best, lewis2019bart, raffel2019exploring} compressing the source language text before generating the target language text. 
Through multiple compressions and generations, back-translation allows to generate paraphrases.
This method has already been successfully used for text augmentation  \cite{sennrich2015improving, xie2019unsupervised}. 
We implemented our version of a multi-step back-translation.

Some works also explore augmentation in word embedding space \cite{zhang2018word, giridhara2019study}.
In such a space, texts are represented as a concatenation of continuous word vectors referred to as word embeddings. 
This continuous text representation can then be used to feed NLP models.
Similarly to computer vision where noise can be applied to augment images \cite{jin2015robust} in the continuous pixel space, Zhang and
Yang \cite{zhang2018word}, and Giridhara et al.\cite{giridhara2019study} study the addition of noise in the continuous word embedding space.
Noise or small variations can occur on image pixels due to sensor uncertainty or adversarial attacks \cite{akhtar2018threat}, consequently noise-based augmentation improves model robustness to small variations. 
However, in natural language, the word embedding for a same word in a same context will not vary as it is the output of a deterministic model.
We argue that adding noise to words embeddings may bring discrepancy as augmented word embeddings cannot exist "naturally". 

It is possible to combine different text augmentation methods. 
For instance, Giridhara et al. \cite{giridhara2019study} combine augmentation at word level and word embedding level, and Wei and Zou \cite{wei2019eda} combine different simple text editing methods.
As in computer vision, we argue that combining different augmentation methods helps to improve augmented texts diversity. 
Our proposed framework uses an ensemble of augmentation methods combining words replacements and back translations (see Section~\ref{sec:augmentation_methods}).

%% file: method.tex
\section{Method}
\label{sec:method}

We propose a new data augmentation framework that combines different methods while ensuring that augmented texts are good candidates to improve the original dataset and the generalization of machine learning models trained on it.

Our proposed augmentation framework generates an augmentation pipeline for each augmentation attempt.
An augmentation pipeline consists of a normalization module that tokenizes the input text, an augmentation module made of a random sequence of $k$ methods that augments the text and a validation module that filters out the augmented text candidate.
At the end of the pipeline the augmented text may be kept or discarded.
We parameterize the number of generated augmentation pipelines for each input text with a constant (e.g., 10). Consequently, the number of kept augmented texts per input is limited and may vary.

This architecture is flexible as we can easily tune, activate, or deactivate the different modules.
We now describe each module in details.

\subsection{Normalization module}
\label{subsec:normalization}
Normalization is the first step of an augmentation pipeline. It parses an input text into separate tokens representing words or punctuation and ensures correct encoding for those tokens.
Our normalization module relies on spaCy\footnote{https://spacy.io/, an open source library for NLP in python and cython.} and provides characteristics about each identified token (e.g., Part-Of-Speech (POS) tag, Named Entities Recognition, stemming). These characteristics are leveraged by our augmentation methods.

\subsection{Augmentation module}
\label{sec:augmentation_methods}
We propose to combine several methods including multi-steps machine translation, words replacement leveraging BERT \cite{devlin2018bert} and Wordnet, abbreviations and misspellings generation based on heuristics.

The number $k$ of methods used in the augmentation module of a pipeline is a parameter (e.g., a random variable in our case, a constant, a function of text length, etc.).
Thus a sequence of $k$ augmentation methods is built picking methods one by one at random with replacement.
Augmentation methods are then applied successively in the sequence order transforming the input normalized text into an augmented text.

\subsubsection{Words replacements:} 
\label{sec:wordrep}

The goal of words replacements is to modify an existing text while preserving its syntax and meaning.
Doing so, we reduce the risk for an augmented text to be grammatically incorrect and incompatible with its label.
We now describe three types of words replacements that we implemented in our framework.

\paragraph{Semantic-based replacements:} With the objective of preserving the original text meaning, we want to replace words by synonyms or near-synonyms \cite{stanojevic2009cognitive}. Leveraging pretrained \textit{BertForMaskedLM base uncased} from  HuggingFace's package  \cite{devlin2018bert}, we developed two different methods to increase replacements diversity. These methods make use of both spaCy and BERT WordPiece tokens as we developed a mapping between these two tokenization techniques.

For the first method, we randomly mask 10\% of WordPiece tokens excluding tokens with spaCy POS tags corresponding to pronouns, determinants and proper nouns.
We then use pretrained BERT to output a probability distribution over the vocabulary for each masked word.  
By masking less than 10\% of tokens, most of the text context is preserved which mitigates the risk of inaccurate probability predictions.
Eventually we choose the replacement candidate sampling over the distribution due to the inherent uncertainty of masked word prediction, leveraging Nucleus Sampling with a threshold of 95\% \cite{holtzman2019curious}.
To mitigate the risk of inconsistent replacements, we added constraints on syntax and contextual semantic continuity. Thus a replacement is accepted if it preserves the original word POS tag (e.g., strict equality) and if the cosine similarity between its embedding and the original word embedding is high enough, i.e., above a given threshold.

For the second method, we use both WordNet \cite{miller1998wordnet} and BERT.
We first perform a selection of words to replace, exactly as done in the first method.
For each selected word, we obtain a list of replacement candidates using WordNet synsets and the word POS tag.
As previously, we compare candidates and original word embeddings with cosine similarity, refining the list. 
Eventually a replacement is randomly chosen from the refined list.


As mentioned in Section~\ref{sec:sota}, other works use language models for words replacements in text augmentation context \cite{kobayashi2018contextual, wu2019conditional, kumar2020data}. 
However, these methods are not corpus agnostic as language models are finetuned, conditioning replacements on text labels. 
Our methods do not involve any finetuning and can directly be used on any text corpus.

\paragraph{Abbreviation-based replacements}: 
Abbreviations are quite common in natural language, so a word or group of words may be replaced by an abbreviation and vice versa (e.g., ``IMO'' for ``In my opinion'').
We perform replacements from expanded to abbreviated form and inversely, relying on word-pair dictionaries.
To ensure that our method remains corpus independent, and to preserve texts meaning, we exclude from those dictionaries abbreviations whose expanded forms differ depending on the texts context (e.g., ``PM'' $\rightarrow$ \{``Post Meridiem'', ``Project Manager'', etc.\}.

\paragraph{Misspelling-based replacements}: 
Misspellings are quite common in natural language and they are often accidental.
We replace a word by a misspelled version relying on a heuristic to simplify misspellings occurrence and letter replacements.
This heuristic can be used on any English text and is detailed in Appendix~\ref{sec:Misspelling replacements}. 
Moreover, in fraud detection context, misspellings are important as they can be used to convey a sense of urgency, or to evade security technologies \cite{howard2008modern, Warner:2012:DHS:2390374.2390377}.

\subsubsection{Multiple steps of translation:}
\label{sec:backtrans}

We propose a method relying on multiple steps of translation using a strongly connected graph. An example of such a graph is given in Appendix~\ref{sec:translation_graph} - Figure~\ref{fig:translationGraph}.
Each node $L_i$ represents a language, and each edge $MT_{ij}$ represents a machine translation engine able to translate from $L_i$ to $L_j$.
We generate a cycle from the $English$ node and passing through several intermediate languages, achieving a multi-steps back-translation.
Numerous constraints can be applied to generate a cycle (e.g., its length, its edge's performance) affecting back-translated texts validity (e.g., label compatibility) and diversity.

In our settings, we use Google Translate engine\footnote{https://cloud.google.com/translate} (GT) for all edges, considering only the best performing translation pairs from and to the English language according to the empirical evaluation available at teachyoubackwards.com\footnote{https://www.teachyoubackwards.com/empirical-evaluation/. The list of languages is provided in Appendix.}.
We also limit the cycle maximum length to 3 (i.e., simple and 2-steps back-translation possible) in our experiments. 
For instance, a path may be: $English \rightarrow  Spanish \rightarrow  Danish \rightarrow English$. 
As mentioned in teachyoubackwards evaluation, it is likely  that GT does not use a direct translation for non-English pairs (e.g., $Spanish \rightarrow  Danish$), but instead uses English language as a pivot (e.g., $Spanish \rightarrow English \rightarrow Danish$).
Leveraging multi-step back-translation, we push further towards back-translated texts diversity while mitigating the risk of invalidity selecting best English translation pairs.

\subsection{Validation module}
\label{subsec:validation}
Ideally an augmentation should ensure label compatibility and bring added value to the dataset.
Making the assumption that diverse paraphrases satisfy these objectives, our validation module goal is to discard augmented texts that are too dissimilar or redundant when compared to the original text.
For this purpose, our validation module consists of two steps: a first step where dissimilar texts are identified by a classifier and a second step where redundant texts are detected by an heuristic.

\subsubsection{Dissimilar text detection:} To discard dissimilar augmented texts, we train a text pair logistic regression classifier on paraphrases corpora to predict if two texts have the same meaning (i.e., \textit{similar}) or not (i.e., \textit{dissimilar}).
Corpora used are Quora Question Pairs \cite{chen2018quora}, Microsoft Research Paraphrases Corpus \cite{dolan2005automatically} and an internal email text pairs dataset (see Appendix~\ref{sec:pairs_label}).
An augmented text is kept and considered for redundancy detection only if it is classified as \textit{similar} to the original text, otherwise it is discarded.

Each input text pair is mapped to a feature vector using a set of similarity metrics.
%
We initially performed a performance study of metrics different enough to increase their potential synergy, considering metrics based on n-grams co-occurrence \cite{papineni2002bleu, huang2008similarity}, metrics leveraging WordNet \cite{mihalcea2006corpus, li2006sentence} or language models \cite{conneau2017supervised, reimers2019sentence}.
This study highlighted a limited complementarity between metrics as most of accuracy is captured by Sentence-bert \cite{reimers2019sentence}.
Thus, in our framework, we combine 3 metrics \cite{mihalcea2006corpus, papineni2002bleu, reimers2019sentence}, slightly improving over Sentence-bert accuracy alone without impacting the computation time.

\subsubsection{Redundant text detection:}
To detect redundancy, we rely on a heuristic, scoring each \textit{similar} augmented text based on the transformations applied (e.g., +1 for each word change). 
Each score is re-scaled based on the original text length (e.g., number of tokens). Example of a score computation is given in Appendix~\ref{sec:augmentation_scoring}.
A \textit{similar} augmented text is considered redundant and discarded if its re-scaled score is too low, i.e., under a given threshold.

To tune this threshold, we rely on our internal text pairs dataset.
As detailed in Appendix~\ref{sec:pairs_label}, each text pair is labeled according to 4 levels of similarity, going from near-duplicate to dissimilar. 
Using the described heuristic, we compute for each pair the re-scaled score comparing one text to another.
Ranking text pairs with re-scaled scores, the threshold optimizes the $F_{1}$ score associated with near-duplicate detection among pairs classified as \textit{similar} by the previously described classifier.

%% file: experiments.tex
\section{Experiments}
\label{sec:experiments}
In this section, we evaluate the contribution of our augmentations to classification tasks commonly used in text augmentation state-of-the-art.
Additionally, we study our augmentations added value in BEC detection use case with different settings, varying the number of augmented texts or using a finetuned language model for better classification performance.

\subsection{Datasets}
\label{subsec:datasets}
To perform our experiments, we use SST-2 \cite{socher2013recursive} and TREC-6 \cite{li2002learning} public datasets as well as our private dataset for BEC detection.
SST-2 is made of movie reviews with two labels (\textit{positive} or \textit{negative}). 
TREC-6 is a multi-label questions dataset, each label corresponding to a question type (e.g., \textit{Abbreviation}, \textit{Locations}, \textit{Entities}).

Our BEC detection dataset is composed of anonymized texts extracted from emails, labeled as suspicious through different typologies (e.g., \textit{ceofraud}, \textit{w2fraud}) or \textit{nonsuspicious} (see Appendix~\ref{sec:bec_labels}).
This dataset is highly unbalanced as BEC are not common.
Class imbalance is a known issue in Machine Learning \cite{krawczyk2016learning} as most models assume data to be equally distributed which results in poor performances on minority classes.
To address this imbalance, we only augment minority classes (i.e., BEC typologies) samples referred as BEC* in Table~\ref{table:datasets_stat}. 

\begin{table}[htb]
\begin{center}
\begin{tabular}{c||c|c|c|c}
\hline
\textbf{Dataset}	 & $c$	& $l$	&  $N$	& $|V|$ \\ 
\hline
SST-2                 & 2	& 19	&  9 613	& 16 185\\ 
\hline
TREC-6                 & 6	& 10	&  5 952 	& 9 592\\ 
\hline
BEC 				 & 6	& 48	&  1 224 	& 8 042 \\
BEC* 				 & 5	& 38	&   115		& 1 153\\
\hline
\end{tabular}
\caption{Datasets statistics}
\label{table:datasets_stat}
\end{center}
\end{table}
Table~\ref{table:datasets_stat} contains for each dataset the number of classes $c$, the average text length\footnote{Average number of words per text rounded to the nearest integer.} $l$, the total number of texts $N$, and the vocabulary size $|V|$.
These datasets are diverse in terms of samples number and text length allowing us to study the impact of these factors on the added value of our augmentations.
Intuitively we expect a stronger added value on a small dataset with long texts.
Indeed, the space of possible augmentations for a text is constrained by its length.
Moreover, a model trained on a large-scale dataset may achieve good generalization without data augmentation.

Besides the number of samples and the text length, the complexity of the classification task must be taken in account.
As mentioned in Section~\ref{sec:sota}, augmented texts must be compatible with their labels to be relevant. 
If the natural language task is complex, augmented texts may hardly ensure label compatibility which would reduce augmentation added value. 
We argue that we also have diversity in term of task complexity:
TREC-6 questions are divided into semantic categories, questions are short and objective;
SST-2 implies figurative language;
BEC classification is sensitive to syntax and semantic as a suspicious email will contain a request on some specific topics (e.g., payment, sharing of sensitive information, etc.).
Thus we analyze the resilience of our augmentation framework to different kinds of classification tasks and complexities.

We only augment the train split of each dataset.
For the public classification tasks, we use TREC-6 and SST-2 provided train/test splits. 
For our BEC detection task, we apply a 70\%/30\% stratified split.

\subsection{Models}
\label{subsec:models}
To study our augmentation contribution on different text classification tasks, we use the \textit{CNN for sentence classification} \cite{kim2014convolutional}.
This model has smaller capacity\footnote{Measure of the complexity, richness, flexibility of the space of functions that can be learned by a machine learning model.} than recent language models \cite{devlin2018bert, yang2019xlnet}, however it has been widely used in text augmentation state-of-the-art \cite{zhang2018word, kobayashi2018contextual, wu2019conditional}. 
%
For the BEC dataset, we also assess our augmentation contribution finetuning \textit{BertForSequenceClassification base uncased} implemented in HuggingFace's \cite{wolf2019huggingface}.
Following Kim \cite{kim2014convolutional} we train our classification models using early stopping on a development set (\textit{dev set}). 
For SST-2 the \textit{dev set} is provided, for TREC-6 and BEC detection datasets we generate it with respectively a 90\%/10\%  and 80\%/20\% stratified split of the train set.

\subsection{Data augmentation}
We augment each dataset using our framework.
As mentioned in Section~\ref{sec:method}, the number of kept augmented texts per input varies.
We generate 10 augmentation pipelines for each text sample, which gives enough augmented texts for our experiments.
The average number of augmented texts per sample (i.e., the augmentation factor) is recorded in Table~\ref{table:augm_fact}.

We argue that the difference between augmentation factors is driven by the average text length (see $l$ in Table~\ref{table:datasets_stat}).
Consequently, the augmentation contribution potential is higher on BEC than on SST-2 or TREC-6.

\begin{table}[htb]
\centering
\begin{tabular}{c||c}
\hline
\textbf{Dataset}    & Augmentation factor \\ \hline
TREC-6 & 7.2 \\ 
SST-2 & 8.8 \\ 
BEC* &  9.2 \\ \hline 
\end{tabular}
\captionsetup{justification=centering}
\caption{Average augmentation factor per dataset.}
\label{table:augm_fact}
\end{table}

\subsection{Impact of data size}

\begin{figure}[!htb]
	\centering
	\captionsetup{justification=centering}
   \resizebox{\linewidth}{!}{
    	\subfloat[TREC-6]{
			\begin{tikzpicture}
			\begin{axis}[
						xlabel=\textit{Originals} proportion,
						ylabel=Accuracy, 
						legend pos=south east,
						grid=both,
						grid style={dotted, line width=.1pt, draw=gray!10},
						minor grid style={line width=.2pt,draw=gray!50},
						major grid style={line width=.2pt,draw=gray!50},
						xmin=0.1,xmax=1,
        				ymin=0.65,ymax=0.9,
						xtick={0.1,0.2,0.3,...,1.1},
						minor tick num=1,
						enlarge x limits=0.1,
						enlarge y limits=0.025
						]
			\addplot [color=black,mark=*] table [
    			x index=0,
    			y index=1] {trec.original};
			\addplot [color=blue,mark=triangle*] table [
    			x index=0,
    			y index=1] {trec.augmentations};
			\addplot [color=green!50!black,mark=square*] table [
    			x index=0,
    			y index=1] {trec.merged};
			\errorband[black, opacity=0.2]{trec.original}{x}{y}{error};
			\errorband[blue, opacity=0.2]{trec.augmentations}{x}{y}{error};
			\errorband[green!50!black, opacity=0.2]{trec.merged}{x}{y}{error};
			\addlegendentry{\textit{Originals}}
			\addlegendentry{\textit{Augmentations}}
			\addlegendentry{\textit{Merged}}
		\end{axis}
			\end{tikzpicture}
		}
		\subfloat[SST-2]{
			\begin{tikzpicture}
			\begin{axis}[
						xlabel=\textit{Originals} proportion,
						legend pos=south east,
						grid=both,
						grid style={dotted, line width=.1pt, draw=gray!10},
						minor grid style={line width=.2pt,draw=gray!50},
						major grid style={line width=.2pt,draw=gray!50},
						xmin=0.1,xmax=1,
        				ymin=0.6,ymax=0.8,
						minor tick num=1,
						xtick={0.1,0.2,0.3,...,1.1},
						enlarge x limits=0.1,
						enlarge y limits=0.05
						]
			\addplot [color=black,mark=*] table [
    			x index=0,
    			y index=1] {sst.original};
			\addplot [color=blue,mark=triangle*] table [
    			x index=0,
    			y index=1] {sst.augmentations};
			\addplot [color=green!50!black,mark=square*] table [
    			x index=0,
    			y index=1] {sst.merged};
			\errorband[black, opacity=0.2]{sst.original}{x}{y}{error};
			\errorband[blue, opacity=0.2]{sst.augmentations}{x}{y}{error};
			\errorband[green!50!black, opacity=0.2]{sst.merged}{x}{y}{error};
			\addlegendentry{\textit{Originals}}
			\addlegendentry{\textit{Augmentations}}
			\addlegendentry{\textit{Merged}}
		\end{axis}
		\end{tikzpicture}
		}
   }
	\caption{Performances on SST-2 and TREC-6 as a function of dataset size. Points show average results, bands the standard deviation considering different model initialization and data downsampling seeds.}
	\label{fig:size_impact}
\end{figure}
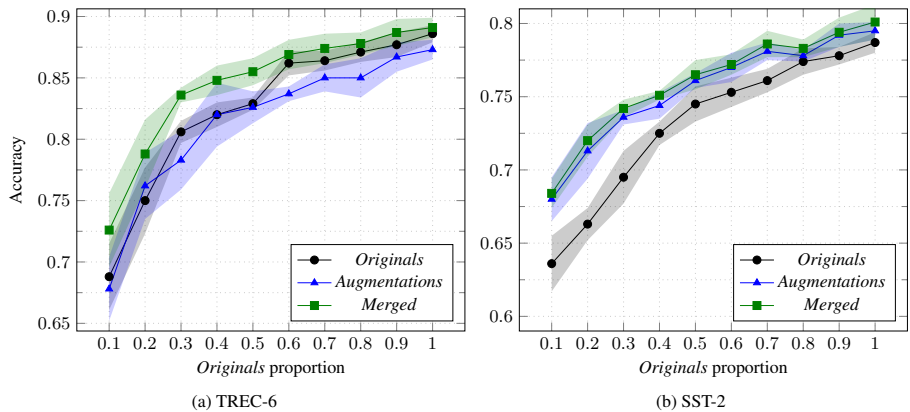

We analyze our augmentation contribution on public datasets through different data regimes. For this purpose, TREC-6 and SST-2 datasets are augmented.
Figure~\ref{fig:size_impact} depicts the results obtained by training the CNN \cite{kim2014convolutional} on three types of data: 
\textbf{\textit{Originals}}, corresponding to text samples from TREC-6 or SST-2 datasets; \textbf{\textit{Augmentations}}, corresponding to augmented texts outputted by our framework using samples from \textit{Originals}; \textbf{\textit{Merged}}, containing all samples from \textit{Originals} and \textit{Augmentations}.
The proportion of \textit{Originals} samples used to create datasets varies from $10\%$ to $100\%$.
For each proportion, a new data trio (\textit{Originals}, \textit{Augmentations}, \textit{Merged}) is generated.
For instance, we may consider a 10\% subsampling of TREC-6, i.e., \textit{Originals}, from this subsample output augmented texts, i.e., \textit{Augmentations},  and combine the subsample and related augmented texts into a single dataset \textit{Merged}.
Evaluation is performed on provided test sets.
Figure~\ref{fig:size_impact} validates that most of our augmented texts preserve label compatibility on different task complexities as there is no significant performance drop when training models solely on augmented samples (i.e., \textit{Augmentations}). 
Figure~\ref{fig:size_impact} also shows the contribution of our augmentation, especially in low-data regimes.
This is depicted by the positive and decreasing performance gap between \textit{Merged} and \textit{Originals}.

\subsection{Impact of augmented data size on an imbalanced corpus} 
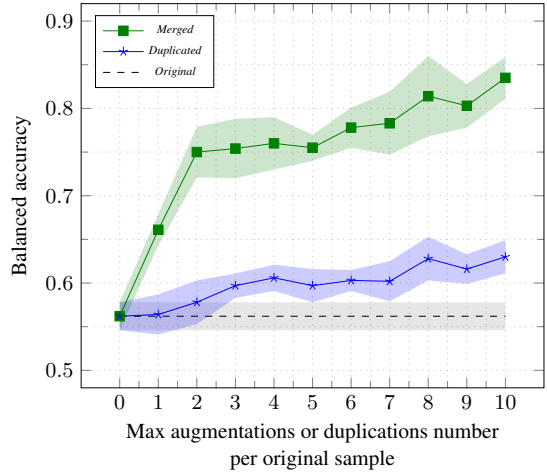
\begin{figure}
	\centering
    \resizebox{0.6\linewidth}{!}{
	\begin{tikzpicture}
		\begin{axis}[
					xlabel style={align=center},
					xlabel=Max augmentations or duplications number\\ per original sample,
					ylabel=Balanced accuracy, 
					legend pos=north west,
					grid=both,
					grid style={dotted, line width=.1pt, draw=gray!10},
					minor grid style={line width=.2pt,draw=gray!50},
					major grid style={line width=.2pt,draw=gray!50},
					minor tick num=1,
					xmin=0,xmax=10,
        			ymin=0.5,ymax=0.9,
					xtick={0,1,2,3,...,10},
					enlarge x limits=.1,
					enlarge y limits=.05
					]
			\addplot [color=green!50!black,mark=square*] table [
    			x index=0,
    			y index=1] {bec.augmented};
			\addplot [color=blue,mark=star] table [
    			x index=0,
    			y index=1] {bec.duplicated};
			\addplot [color=black,mark=none,dashed] table [
    			x index=0,
    			y index=1] {bec.original};
			\errorband[black, opacity=0.1]{bec.original}{x}{y}{error};
			\errorband[blue, opacity=0.2]{bec.duplicated}{x}{y}{error};
			\errorband[green!50!black, opacity=0.2]{bec.augmented}{x}{y}{error};
			\addlegendentry{\textit{\tiny{Merged}}}
			\addlegendentry{\textit{\tiny{Duplicated}}}
			\addlegendentry{\textit{\tiny{Original}}}
		\end{axis}
	\end{tikzpicture}
	}
	\caption{Augmentation factor impact on BEC corpus. Points show average results, bands the standard deviation using different model initialization seeds.}  
	\label{fig:aug_cap}
\end{figure}

To improve our email text classifier generalization for BEC detection, we augment our BEC corpus on minority classes. 
\textit{Merged} is the dataset containing BEC corpus and all related augmented texts. 
On average 9.2 augmented texts are generated for each sample (Table~\ref{table:augm_fact}), which significantly increases minority classes weight in the learning phase. 
To dissociate classes weight effect from our augmentation contribution, we created BEC ``duplicated'' corpus (\textit{Duplicated}) where each minority class sample is duplicated to replicate the number of augmented texts in \textit{Merged}.
To analyze the impact of the number of augmented texts per sample, we capped\footnote{A cap only limits the maximum number of augmented texts per sample to a given value,  it does not constrain the minimum number of augmented texts.} this number from $0$ to $10$ and trained CNN \cite{kim2014convolutional} on each data type (i.e., \textit{Original}, \textit{Duplicated} and \textit{Merged}), results are shown in Figure~\ref{fig:aug_cap}. 

Augmentation contribution is significant in this small-scale dataset (see Table~\ref{table:datasets_stat}) as we can notice a large gap between \textit{Merged} and other data types.
Moreover, the contribution increases with the cap, first strongly then moderately after a cap of 2 augmented texts per sample.
The performance growth shows our framework ability to produce both diverse and label compatible augmented texts.
However, the growth slowdown may indicate that diversity among augmented texts becomes harder to maintain as their number increases. 
Yet, it may also be explained by other factors (e.g., model capacity). For this reason, we analyzed text samples with their related augmentations.
This analysis (examples in Appendix~\ref{sec:aug_redundancy}) shows that diversity among augmented texts can further be improved.

\subsection{Our use case with class imbalance}
\label{subsec:bec_det}
\begin{table}[htb]
\begin{center}
\begin{tabular}{c||c|c}
\hline
\textbf{Dataset}	& \thead{CNN \cite{kim2014convolutional}}	& \thead{BertForSequence \\ Classification}	 \\ \hline
Original            & 0.562 & 0.903 \\ 
Duplicated    		& 0.619 & 0.912	\\ 
Augmented 	  		& 0.834 & 0.960	\\ \hline
\end{tabular}
\caption{BEC detection balanced accuracy}
\label{table:bec_bal_acc}
\end{center}
\end{table}
In this experiment, \textit{BertForSequenceClassification} is considered in addition to CNN \cite{kim2014convolutional}.
As mentioned previously, language models have improved NLP state-of-the-art on various tasks\footnote{https://gluebenchmark.com/leaderboard}. 
Leveraging self-supervised pretraining and Transformer architecture \cite{vaswani2017attention}, these models learn and transpose language concepts to specific tasks, achieving excellent generalization.

Table~\ref{table:bec_bal_acc} regroups the best results obtained when varying the cap of the augmented texts number per sample.
As shown previously (Figure~\ref{fig:aug_cap}), augmentation is key for CNN model generalization, with a contribution well above classes weight.
With \textit{BertForSequenceClassification}, while being less significant, augmentation still has a positive impact over class weights.
Thus we exemplified that even with recent pretrained language models such as BERT, data augmentation stays relevant in low-data regimes.

%% file: arg.tex
\section{Further analysis}
\label{sec:arg}

\subsection{Corpus agnostic framework}
We designed our augmentation framework to be corpus agnostic in order to address the diversity of use cases within our company.
Initially motivated by the task of BEC detection, the development of our framework evolved towards an internal service to augment English texts in several contexts (e.g., BEC detection, security awareness training, fraud simulation).
Being corpus agnostic may degrade the quality of augmented texts as it prevents the use of context specific methods, such as model finetuning \cite{kobayashi2018contextual, wu2019conditional}.
We propose several workarounds to mitigate this risk.

First, as detailed in Section~\ref{sec:method}, we constrain words replacements.
Replacement candidates must preserve their initial POS tags and are filtered based on the cosine similarity of BERT embeddings.
These constraints reduce inconsistent replacements. However they are not optimal as BERT embeddings are not optimized for cosine similarity measurements which suppose that all dimensions have equal weights\footnote{https://github.com/UKPLab/sentence-transformers/issues/80}, and the POS tag preservation constraint may discard some valid replacements.
 
Additionally, we developed a validation module called at the end of the pipeline (Section~\ref{sec:method}) to reduce the risk of compromising labels.
This module is further discussed in following subsections.
Results presented in Section~\ref{sec:experiments} have shown, in different contexts, that our framework produces label compatible augmentations despite being corpus agnostic.

\subsection{Augmentations and diversity} 

\begin{figure}[h]
\centering
\includegraphics[width=.6\linewidth]{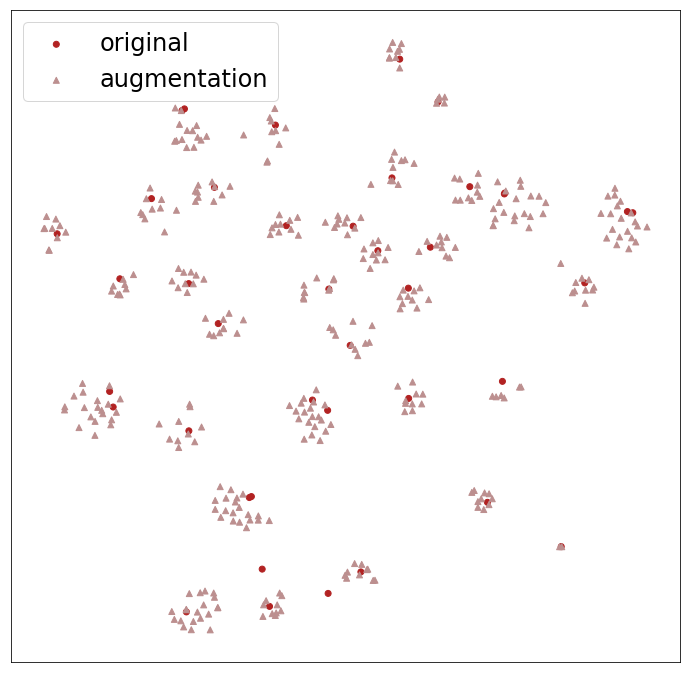}
\captionsetup{justification=centering}
\caption{2D visualization.}
\label{fig:2d_rep}
\end{figure}

To ease the analysis of augmentation impact on data diversity, we propose a visualization of augmented \textit{areyouavailable} class from BEC corpus (Figure~\ref{fig:2d_rep}), projecting each text to an embedding space \cite{reimers2019sentence} and leveraging t-SNE \cite{maaten2008visualizing} for 2D visualization\footnote{2D visualization implies a loss of information about data representation due to the significant dimension reduction.}.

Visualization suggests that augmented texts, while bringing diversity, are restricted to the neighborhood of original samples they relate to.
The challenge of diversity is to expand these neighborhoods while preserving augmented texts validity.
We argue that using an ensemble of augmentation methods helps increase diversity.
Our experiments (Section~\ref{sec:experiments}) exhibit the ability and limitations of our framework to continuously produce diverse augmented texts.
Diversity improvement will be addressed in future work by modifying our existing methods (e.g., translation paths addition, words replacements sampling change) and investigating new ones involving for instance natural language generation.

\subsection{Augmentations validation}
It is important to ensure that augmented texts have enough diversity while preserving data characteristics, which is a challenge as our augmentation framework is corpus agnostic. 
Indeed, we cannot guarantee that our transformations always preserve data characteristics.
For this purpose, we use a validation module constraining our augmented texts to be paraphrases of the samples they relate to. 
We made the assumption that paraphrasing original texts brings sufficient control to prevent data poisoning in most contexts.
Moreover, our validation module also discards detected near duplicates of original texts to augment data diversity (Section~\ref{sec:method}).
However, this validation module is not an oracle. Paraphrase detection models tend to fail on sentence pairs that have lexical overlap but different meaning \cite{zhang2019paws} which can easily occur in augmentation context.
In future work, we will investigate language models finetuning which have shown top performance on paraphrase detection\footnote{https://paperswithcode.com/sota/semantic-textual-similarity-on-mrpc} and consider Paraphrase Adversaries from Word Scrambling corpus \cite{zhang2019paws} as an additional data source.

\subsection{Task influence} 
Similarly to what is done for image or sound augmentation, it is relevant to analyze how data is processed and what we know about the task before designing augmentation.
Our framework being corpus and task agnostic, this preliminary analysis is not possible.
Moreover, constraining augmented texts to be valid paraphrases of original texts may not prevent data poisoning for all tasks.
For example, if the task is authorship attribution (i.e. determining the writer of a text), using back-translation to augment texts may poison the data, altering some writer style characteristics (e.g., vocabulary distribution, syntax representation).
The modular design of our framework brings flexibility helping to cope with some specific contexts, for instance we may not consider multi-step back-translation in case of authorship attribution.

%% file: conclusion.tex
\section{Closing remarks}
In this paper we present a new text augmentation framework that is corpus and task agnostic. 
This framework combines several augmentation methods and can be easily applied to different NLP projects.
Experiments have shown that our augmentation improves significantly machine learning models generalization in different low-data regime contexts.
Our analysis on limitations provides guidelines for future work such as the addition of new methods (e.g., natural language generation) or the improvement of existing ones (e.g., replacements monitoring, validation module) aiming towards better augmented texts diversity and validity.

%% file: acknowledgments.tex
\section*{Acknowledgments}

We are grateful to Tatiana Ekeinhor, Thomas Gendron, Antoine Honor\'{e}, Paolo Pinto, and Damien Riquet for their reviews and involvement in the 
augmentation framework development.
Finally we would like to acknowledge Fatima Qachfar's contribution to obtain early results on the use of language models.

%% file: appendix.tex
\newpage

\section{Misspelling replacements heuristic}
\label{sec:Misspelling replacements}
Below is the code of the main function behind our misspelling replacements heuristic. A token has a probability of 0.01 to be picked for misspelling replacement.

\begin{lstlisting}
def generate_misspellings(picked_token):
    word = picked_token.string

    # default misspelling types
    misspelling_types = ["transposition", "extra", "mistyping", "phonetic"]

    # optional misspelling type 
    has_doubles = any(word[i] == word[i + 1] for i in range(len(word) - 1))
    if has_doubles:
        misspelling_types.append("missing")

    # pick misspelling
    misspelling_type = random.choice(misspelling_types)
    misspelled_word = apply_misspelling(word, misspelling_type)
    picked_token.string = misspelled_word
    return picked_token
\end{lstlisting}

The table below described \textit{apply\_misspelling()} action for each misspelling type:

\begin{table*}
\centering
\begin{tabularx}{\textwidth}{c|c|c|X}
\hline
\textbf{Input token}	 & \textbf{Misspelling type}	& \textbf{Output token}	&  \textbf{Description} \\ 
\hline
winter                & transpostion	& witner	&  Swap two adjacent letters	\\ 
\hline
winter                & extra	& winterr	&  Repeat a letter twice	\\ 
\hline
winter 				 & mistyping	& wimter	&  Replace a letter by an adjacent letter in QWERTY layout 	\\
\hline
misspelling 		& missing	& misspeling	&   Remove a letter in case of double letters		\\
\hline
dependent 			& phonetic	& dependant	&   Replace a group of letters by another which has the same pronunciation	\\
\hline
\end{tabularx}
\caption{\textit{apply\_misspelling()} actions}
\end{table*}

\section{Example of multistep translation graph}
\label{sec:translation_graph}
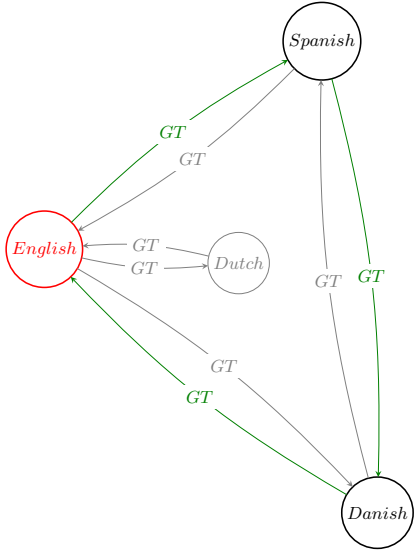
\begin{figure}[ht]
	\centering     
 	\resizebox{0.45\linewidth}{!}{
  		\centering 
  		\begin{tikzpicture}

  \tikzset{VertexStyle/.style = {shape          = circle,
  								 black!50,
                                 draw,
                                 text           = black!50,
                                 inner sep      = 2pt,
                                 outer sep      = 0pt,
                                 minimum size   = 24 pt}}
  \tikzset{EdgeStyle/.style   = {black!50,
  								  -stealth}}
  \tikzset{LabelStyle/.style =   {
                                  fill           = white,
                                  text           = black!50}}
    \node[VertexStyle, thick, red, text = red] (A) at (0,5.25) {\textbf{$English$}};
    \node[VertexStyle, thick, black, text = black] (B) at (5,9) {$Spanish$};
    \node[VertexStyle, thick, black, text = black] (C) at (6,0.5) {$Danish$};
    \node[VertexStyle] (D) at (3.5,5) {$Dutch$};

    \tikzset{EdgeStyle/.append style = {bend left=8}}
	\draw[EdgeStyle, black!50!green](A) to node[LabelStyle, text = black!50!green]{$GT$} (B);
	\draw[EdgeStyle](B) to node[LabelStyle]{$GT$} (A);
	
	\tikzset{EdgeStyle/.append style = {bend left=8}}	
	\draw[EdgeStyle](A) to node[LabelStyle]{$GT$} (C);
	\draw[EdgeStyle, black!50!green](C) to node[LabelStyle, text = black!50!green]{$GT$} (A);
	
	\tikzset{EdgeStyle/.append style = {bend right=9}}
	\draw[EdgeStyle](A) to node[LabelStyle]{$GT$} (D);
	\draw[EdgeStyle](D) to node[LabelStyle]{$GT$} (A);
	
	\tikzset{EdgeStyle/.append style = {bend left=8}}
	\draw[EdgeStyle](C) to node[LabelStyle]{$GT$} (B);
	\draw[EdgeStyle, black!50!green](B) to node[LabelStyle, text = black!50!green]{$GT$} (C);

  \end{tikzpicture}}
  \caption{Example of translation graph}
  \label{fig:translationGraph}
\end{figure}

Based on teachyoubackwards\footnote{https://www.teachyoubackwards.com/empirical-evaluation/} study, we chose the following list of languages: Afrikaans, Chinese, Croatian, Danish, Dutch, Finnish, German, Greek, Hungarian, Polish, Portuguese, Spanish. All these languages show relatively good translation performances from and to English, i.e, $BARD\geq40$, $TARZAN\geq50$, $FAIL\leq35$.
On Figure~\ref{fig:translationGraph}, the following set of languages $\{English, Dutch, Danish, Spanish\}$ is considered.
$GT$ which stands for Google Translate engine is used for all translations.
For each back-translation, a cycle is generated from $English$ node. In the example represented on Figure~\ref{fig:translationGraph}, the generated path is 
: $English \rightarrow  Spanish \rightarrow  Danish \rightarrow English$.

\section{Example of augmentation scoring}
\label{sec:augmentation_scoring}
In Table~\ref{fig:augmentation_scoring}, we show an example of score computation for the redundancy detection heuristic. In this case, the re-scaling is done by dividing the score by the number of SpaCy tokens in the original text.

\begin{table*}
\centering
\begin{tabularx}{\textwidth}{c|X|X|c|c}
\hline
\textbf{Step} & \textbf{Text} & \textbf{Transformation}   & \textbf{Score} & \textbf{Rescaled}
\\
\hline
0 & ``Kindly pay the attached invoice today and forward remittance once paid.'' & NA / Original text & $s \leftarrow 0$ & $\frac{0}{12}= 0$
\\
\hline
1 & ``Pay the enclosed invoice today and send it in when it's paid.'' & Translation: 5 words changed in input \textit{(Kindly, attached, forward, remittance, once)}  & $s \leftarrow s + 5$ & $\frac{5}{12} \approx 0.42$
\\
\hline
2 & ``Pay the enclosed invoice ttoday and send it in when it's paid.'' & Misspelling: 1 word changed in input \textit{(today)} & $s \leftarrow s + 1$ & $\frac{6}{12}= 0.50$
\\
\hline
\end{tabularx}
\caption{Augmentation scoring}
\label{fig:augmentation_scoring}
\end{table*}

\section{BEC corpus labels}
\label{sec:bec_labels}
Table~\ref{table:bec_labels} describes BEC corpus classes.

\begin{table*}
\centering
\begin{tabularx}{\textwidth}{c|X|X}
\hline
\textbf{Label} & \textbf{Definition}   & \textbf{Example} 
\\
\hline
\textit{areyouavailable} & Short email asking for the availability of the recipient, usually with a sense of urgency. One of the goal of this kind of email is to initiate a communication with the victim without using suspect keywords. BEC detection technologies may whitelist an email address after a couple of emails have been exchanged, hence this trick. &  ``I need you to get a task done for me promptly. Let me know if you are unoccupied''
\\
\hline
\textit{ceofraud} & CEO fraud is a threat consisting in the impersonation  of a position of authority (e.g., CEO) asking an employee to transfer money to a fraudulent bank account. It usually targets finance department. The threat relies on the fear of failing to please an authoritative power. & ``Please process immediately a wire transfer payment to \$45,000.It is an urgent invoice from the business attorney. Banking instructions attached. Thanks'' 
\\
\hline
\textit{giftcardscam} & Gift card scam is a threat consisting in the impersonation of an executive requesting the purchase of gift cards for a special occasion. The message will often tell the victim to stay quiet, trying to make it appear as a surprise. & ``Hi, I have a request, I'm planning to surprise some of the staff with gifts, are you available to get a purchase done for me? I will appreciate your assistance and confidentiality. Email me once you get this.'' 
\\
\hline
\textit{payrollfraud} & Payroll fraud is a threat consisting in the impersonation of an employee asking to change direct deposit information. It usually targets HR or finance departments. The goal for the fraudster is to replace in the payroll system the impersonated employee bank account by a fraudulent bank account. &  ``I have recently changed banks and need to change my direct deposit for payroll. I need your prompt assistance in this matter and when this will take effect.''
 \\
\hline
\textit{w2fraud} & W-2 fraud is a threat consisting in the impersonation of an executive asking for documents (e.g., W-2 forms) containing employees confidential information, such as SSN. & ``I have an important meeting at 10:30 and I need the W-2s of the staff ASAP. Please upload the files to https://drop.box.it for security purposes. Thanks!'' 
\\
\hline
\textit{nonsuspicious} & This label corresponds to all emails we decided not to flag in our BEC detection context, i.e., emails that do not belong to previous labels. There are two types of emails in this category: benign emails that have similar topics  (invoice, sense of urgency, etc.), and benign emails that do not have any similar topics (majority of the corpus).
&  ``I will pay this invoice per Randy.  If you need anything  else just let me know. Thanks.'' 
\\
\hline
\end{tabularx}
\caption{BEC corpus labels definition}
\label{table:bec_labels}
\end{table*}

\section{Redundancy among augmentations}
\label{sec:aug_redundancy}
An analysis of augmented texts relative to randomly selected original texts pointed out a limited diversity. Table~\ref{table:aug_red} gives examples of very similar augmented texts.

\begin{table*}
\centering
\begin{tabularx}{\textwidth}{c|X|X|X}
\hline
\textbf{Dataset} & \textbf{Original sample} & \textbf{Augmentation\SB{1}}   & \textbf{Augmentation\SB{2}}
\\
\hline
BEC Corpus & ``I need you to initiate a wire transfer in the sum of \$45,250 to the account below. I am boarding a flight and this needs to be done right now. Can you please get this done? Send confirmation of the transfer immediately. Thanks'' & ``I want you to initiate a 54 300 USD wire transfer to the ACCT below. I'm boarding the plane and I have to do it now. Can you do this? Send the transfer confirmation immediately. Thanks''&  ``I want you to initiate a transfer of 31,987 EUR to the account below. I'm boarding the plane and I have to do it now. Can you do this? Send transfer confirmation immediately. Thanks''
\\
\hline
BEC Corpus & ``I have important request i need you to handle immediately. Kindly confirm your availability. Regards'' & ``I have an important request I need you to deal with immediately. Plesae confirm your availability. Regards.'' &  ``I have an important request, you must deal with it immediately. Confirm your availability. Regards'' 
\\
\hline
BEC Corpus &``I need you to get a task done for me promptly. Let me know if you are unoccupied'' & ``I need you to do a job for me right away. Let me know if you are free'' &  ``I need you to do the job for me right away. Please let me know if you are not busy'' 
\\
\hline
SST-2 &``about as exciting to watch as two last place basketball teams playing one another on the final day of the season'' & ``Just as exciting as the last two basketball team on the last day of the season'' &  ``It's as exciting as the last two basketball teams on the last day of the season'' 
\\
\hline
SST-2 &``it will guarantee to have you leaving the theater with a smile on your face'' & ``it ensures you leave the theater with a smile'' &  ``it ensures that you leave the room with a smile'' 
\\
\hline
SST-2 &``it is a likable story , told with competence'' & ``it's a pleasant story, skillfully told'' &  ``it is a pleasant story, sillfully told'' 
\\
\hline
TREC-6 &``What does the double-O indicate in 007 ?'' & ``What does double O 007 mean?'' &  ``What does the doubled O 007 mean?'' 
\\
\hline
TREC-6 &``How can I give myself a French manicure ?'' & ``How do I do a French manicure?'' &  ``How can I do a French manicure?'' 
\\
\hline
\end{tabularx}
\caption{Redundancy among augmentations}
\label{table:aug_red}
\end{table*}

\section{Internal text pairs dataset labels}
\label{sec:pairs_label}
Labels correspond to different levels of similarity considering texts pairs meaning and lexical overlap. We define:
\begin{description}
\item[$\bullet$] \textit{0}: texts have different meaning and low lexical overlap.
\item[$\bullet$] \textit{1}: texts have different meaning and significant lexical overlap.
\item[$\bullet$] \textit{2}: texts have same meaning with low to moderate lexical overlap.
\item[$\bullet$] \textit{3}: texts have same meaning and are nearly identical (i.e., high lexical overlap).
\end{description}

In the context of text pair classification, labels \textit{0} and \textit{1} can be grouped under \textit{dissimilar}, labels \textit{2} and \textit{3} under \textit{similar}.

To build the corpus, we first extracted email texts from our professional mailboxes. These email texts have been manually anonymized.
We also extracted texts from public email corpora such as Enron \cite{klimt2004introducing}, DNC\footnote{https://wikileaks.org/dnc-emails/}, Podesta\footnote{https://wikileaks.org/podesta-emails/}.
Then for each selected text, we handcrafted new texts in order to form pairs for the defined labels. Samples from our corpus are given in Table~\ref{table:internal}.

\begin{table*}
\centering
\begin{tabularx}{\textwidth}{X|X|X|X}
\hline
\textbf{0}    & \textbf{1} & \textbf{2} & \textbf{3}\\
\hline
    (``Can you please send me the SSN of all employees ASAP. Thanks'',
    
     ``Spring is coming, my favorite time of the year.'') & (``Can you please send me the SSN of all employees ASAP. Thanks'',
     
      ``Would you send me the picture we took with all employees?'') & (``Can you please send me the SSN of all employees ASAP. Thanks'',
      
       ``This is urgent, could you send me the social security numbers of the whole staff? Thanks'') & (``Can you please send me the SSN of all employees ASAP. Thanks'',
       
        ``Could you send me the SSN of all employees ASAP. Thank you'') \\
\\
\hline
(``Kindly pay the attached invoice today and forward remittance once paid.'',
    
     ``The shipment will be sent as soon as we receive confirmation.'') & (``Kindly pay the attached invoice today and forward remittance once paid.'',
     
      ``Please reject the attached invoice asap and send me confirmation once done.'') & (``Kindly pay the attached invoice today and forward remittance once paid.'',
      
       ``Please pay the joined invoice by eod and send me the associated receipt.'') & (``Kindly pay the attached invoice today and forward remittance once paid.'',
       
        ``Please pay the attached invoice today and forward me the remittance once paid.'') \\
\\
\hline
(``Are you at work?'',
    
     ``Coming home soon.'') & (``Are you at work?'',
     
      ``Are you at the gym?'') & (``Are you at work?'',
      
       ``Are you in the office?'') & (``Are you at work?'',
       
        ``you're at work?'') \\
\\
\hline
(``I have important request I need you to handle immediately. Kindly confirm your availability.'',
    
     ``I am learning that you are going to transfer?! Why is that? I am kind of sad.'') & (``I have important request I need you to handle immediately. Kindly confirm your availability.'',
     
      ``I am booking the restaurant immediately, confirm your availablility.'') & (``I have important request I need you to handle immediately. Kindly confirm your availability.'',
      
       ``Are you available? There is a very important request I want you to take care of.'') & (``I have important request I need you to handle immediately. Kindly confirm your availability.'',
       
        ``I have an important request I need you to handle immediately. Please confirm availability.'') \\
\\
\hline
\end{tabularx}
\caption{Internal text pair samples}
\label{table:internal}
\end{table*}

%% file: main.bbl
\begin{thebibliography}{10}
\providecommand{\url}[1]{\texttt{#1}}
\providecommand{\urlprefix}{URL }
\providecommand{\doi}[1]{https://doi.org/#1}

\bibitem{akhtar2018threat}
Akhtar, N., Mian, A.: Threat of adversarial attacks on deep learning in
  computer vision: A survey. IEEE Access  \textbf{6},  14410--14430 (2018)

\bibitem{bojar2016findings}
Bojar, O., Chatterjee, R., Federmann, C., Graham, Y., Haddow, B., Huck, M.,
  Yepes, A.J., Koehn, P., Logacheva, V., Monz, C., et~al.: Findings of the 2016
  conference on machine translation. In: Proceedings of the First Conference on
  Machine Translation: Volume 2, Shared Task Papers. pp. 131--198 (2016)

\bibitem{chen2018best}
Chen, M.X., Firat, O., Bapna, A., Johnson, M., Macherey, W., Foster, G., Jones,
  L., Parmar, N., Schuster, M., Chen, Z., et~al.: The best of both worlds:
  Combining recent advances in neural machine translation. arXiv preprint
  arXiv:1804.09849  (2018)

\bibitem{chen2018quora}
Chen, Z., Zhang, H., Zhang, X., Zhao, L.: Quora question pairs (2018)

\bibitem{cidon2019high}
Cidon, A., Gavish, L., Bleier, I., Korshun, N., Schweighauser, M., Tsitkin, A.:
  High precision detection of business email compromise. In: 28th
  $\{$USENIX$\}$ Security Symposium ($\{$USENIX$\}$ Security 19). pp.
  1291--1307 (2019)

\bibitem{conneau2017supervised}
Conneau, A., Kiela, D., Schwenk, H., Barrault, L., Bordes, A.: Supervised
  learning of universal sentence representations from natural language
  inference data. arXiv preprint arXiv:1705.02364  (2017)

\bibitem{devlin2018bert}
Devlin, J., Chang, M.W., Lee, K., Toutanova, K.: Bert: Pre-training of deep
  bidirectional transformers for language understanding. arXiv preprint
  arXiv:1810.04805  (2018)

\bibitem{dolan2005automatically}
Dolan, W.B., Brockett, C.: Automatically constructing a corpus of sentential
  paraphrases. In: Proceedings of the Third International Workshop on
  Paraphrasing (IWP2005) (2005)

\bibitem{edunov2018understanding}
Edunov, S., Ott, M., Auli, M., Grangier, D.: Understanding back-translation at
  scale. arXiv preprint arXiv:1808.09381  (2018)

\bibitem{fbiICR2020}
FBI: 2020 internet crime report. Tech. rep., Federal Bureau of Investigation,
  Internet Crime Complaint Center (03 2021)

\bibitem{giridhara2019study}
Giridhara, P.K.B., Chinmaya, M., Venkataramana, R.K.M., Bukhari, S.S., Dengel,
  A.: A study of various text augmentation techniques for relation
  classification in free text. In: Proceedings of the 8th International
  Conference on Pattern Recognition Applications and Methods (2019)

\bibitem{gupta2018deep}
Gupta, A., Agarwal, A., Singh, P., Rai, P.: A deep generative framework for
  paraphrase generation. In: Thirty-Second AAAI Conference on Artificial
  Intelligence (2018)

\bibitem{ho2017detecting}
Ho, G., Sharma, A., Javed, M., Paxson, V., Wagner, D.: Detecting credential
  spearphishing in enterprise settings. In: 26th $\{$USENIX$\}$ Security
  Symposium ($\{$USENIX$\}$ Security 17). pp. 469--485 (2017)

\bibitem{holtzman2019curious}
Holtzman, A., Buys, J., Du, L., Forbes, M., Choi, Y.: The curious case of
  neural text degeneration. arXiv preprint arXiv:1904.09751  (2019)

\bibitem{howard2008modern}
Howard, F.: Modern web attacks. Network Security  \textbf{2008}(4),  13--15
  (2008)

\bibitem{huang2008similarity}
Huang, A.: Similarity measures for text document clustering. In: Proceedings of
  the sixth new zealand computer science research student conference
  (NZCSRSC2008), Christchurch, New Zealand. vol.~4, pp. 9--56 (2008)

\bibitem{idika2007survey}
Idika, N., Mathur, A.P.: A survey of malware detection techniques. Purdue
  University  \textbf{48},  2007--2 (2007)

\bibitem{jakobsson2016understanding}
Jakobsson, M.: Understanding social engineering based scams. Springer (2016)

\bibitem{jin2015robust}
Jin, J., Dundar, A., Culurciello, E.: Robust convolutional neural networks
  under adversarial noise. arXiv preprint arXiv:1511.06306  (2015)

\bibitem{kim2014convolutional}
Kim, Y.: Convolutional neural networks for sentence classification. arXiv
  preprint arXiv:1408.5882  (2014)

\bibitem{kingma2013autoencoding}
Kingma, D.P., Welling, M.: Auto-encoding variational bayes (2013)

\bibitem{klimt2004introducing}
Klimt, B., Yang, Y.: Introducing the enron corpus. In: CEAS (2004)

\bibitem{kobayashi2018contextual}
Kobayashi, S.: Contextual augmentation: Data augmentation by words with
  paradigmatic relations. arXiv preprint arXiv:1805.06201  (2018)

\bibitem{krawczyk2016learning}
Krawczyk, B.: Learning from imbalanced data: open challenges and future
  directions. Progress in Artificial Intelligence  \textbf{5}(4),  221--232
  (2016)

\bibitem{kumar2020data}
Kumar, V., Choudhary, A., Cho, E.: Data augmentation using pre-trained
  transformer models. arXiv preprint arXiv:2003.02245  (2020)

\bibitem{kwak2020users}
Kwak, Y., Lee, S., Damiano, A., Vishwanath, A.: Why do users not report spear
  phishing emails? Telematics and Informatics p. 101343 (2020)

\bibitem{levow2009multilevel}
Levow, Z., Drako, D.: Multilevel intent analysis method for email filtration
  (Dec~3 2009), uS Patent App. 12/128,286

\bibitem{lewis2019bart}
Lewis, M., Liu, Y., Goyal, N., Ghazvininejad, M., Mohamed, A., Levy, O.,
  Stoyanov, V., Zettlemoyer, L.: Bart: Denoising sequence-to-sequence
  pre-training for natural language generation, translation, and comprehension.
  arXiv preprint arXiv:1910.13461  (2019)

\bibitem{li2002learning}
Li, X., Roth, D.: Learning question classifiers. In: Proceedings of the 19th
  international conference on Computational linguistics-Volume 1. pp.~1--7.
  Association for Computational Linguistics (2002)

\bibitem{li2006sentence}
Li, Y., McLean, D., Bandar, Z.A., O'shea, J.D., Crockett, K.: Sentence
  similarity based on semantic nets and corpus statistics. IEEE transactions on
  knowledge and data engineering  \textbf{18}(8),  1138--1150 (2006)

\bibitem{li2017paraphrase}
Li, Z., Jiang, X., Shang, L., Li, H.: Paraphrase generation with deep
  reinforcement learning. arXiv preprint arXiv:1711.00279  (2017)

\bibitem{maaten2008visualizing}
Maaten, L.v.d., Hinton, G.: Visualizing data using t-sne. Journal of machine
  learning research  \textbf{9}(Nov),  2579--2605 (2008)

\bibitem{mansfield2016imitation}
Mansfield-Devine, S.: The imitation game: How business email compromise scams
  are robbing organisations. Computer Fraud \& Security  \textbf{2016}(11),
  5--10 (2016)

\bibitem{mccarthy1986applications}
McCarthy, J.: Applications of circumscription to formalizing common-sense
  knowledge. Artificial intelligence  \textbf{28}(1),  89--116 (1986)

\bibitem{mihalcea2006corpus}
Mihalcea, R., Corley, C., Strapparava, C., et~al.: Corpus-based and
  knowledge-based measures of text semantic similarity. In: Aaai. vol.~6, pp.
  775--780 (2006)

\bibitem{miller1998wordnet}
Miller, G.A.: WordNet: An electronic lexical database. MIT press (1998)

\bibitem{nissim2015detection}
Nissim, N., Cohen, A., Glezer, C., Elovici, Y.: Detection of malicious pdf
  files and directions for enhancements: A state-of-the art survey. Computers
  \& Security  \textbf{48},  246--266 (2015)

\bibitem{oliver2008fraudulent}
Oliver, J., Eikenberry, S.D., Budman, G., Kim, B.: Fraudulent message detection
  (Nov~11 2008), uS Patent 7,451,487

\bibitem{papineni2002bleu}
Papineni, K., Roukos, S., Ward, T., Zhu, W.J.: Bleu: a method for automatic
  evaluation of machine translation. In: Proceedings of the 40th annual meeting
  on association for computational linguistics. pp. 311--318. Association for
  Computational Linguistics (2002)

\bibitem{perez2017effectiveness}
Perez, L., Wang, J.: The effectiveness of data augmentation in image
  classification using deep learning. arXiv preprint arXiv:1712.04621  (2017)

\bibitem{proofpoint180315}
Proofpoint: Understanding email fraud. Tech. rep. (03 2018)

\bibitem{radford2019language}
Radford, A., Wu, J., Child, R., Luan, D., Amodei, D., Sutskever, I.: Language
  models are unsupervised multitask learners. OpenAI Blog  \textbf{1}(8), ~9
  (2019)

\bibitem{raffel2019exploring}
Raffel, C., Shazeer, N., Roberts, A., Lee, K., Narang, S., Matena, M., Zhou,
  Y., Li, W., Liu, P.J.: Exploring the limits of transfer learning with a
  unified text-to-text transformer. arXiv preprint arXiv:1910.10683  (2019)

\bibitem{reimers2019sentence}
Reimers, N., Gurevych, I.: Sentence-bert: Sentence embeddings using siamese
  bert-networks. arXiv preprint arXiv:1908.10084  (2019)

\bibitem{rezende2014stochastic}
Rezende, D.J., Mohamed, S., Wierstra, D.: Stochastic backpropagation and
  approximate inference in deep generative models (2014)

\bibitem{salamon2017deep}
Salamon, J., Bello, J.P.: Deep convolutional neural networks and data
  augmentation for environmental sound classification. IEEE Signal Processing
  Letters  \textbf{24}(3),  279--283 (2017)

\bibitem{sennrich2015improving}
Sennrich, R., Haddow, B., Birch, A.: Improving neural machine translation
  models with monolingual data. arXiv preprint arXiv:1511.06709  (2015)

\bibitem{socher2013recursive}
Socher, R., Perelygin, A., Wu, J., Chuang, J., Manning, C.D., Ng, A.Y., Potts,
  C.: Recursive deep models for semantic compositionality over a sentiment
  treebank. In: Proceedings of the 2013 conference on empirical methods in
  natural language processing. pp. 1631--1642 (2013)

\bibitem{stanojevic2009cognitive}
Stanojevi{\'c}, M., et~al.: Cognitive synonymy: A general overview. FACTA
  UNIVERSITATIS-Linguistics and Literature  \textbf{7}(2),  193--200 (2009)

\bibitem{vaswani2017attention}
Vaswani, A., Shazeer, N., Parmar, N., Uszkoreit, J., Jones, L., Gomez, A.N.,
  Kaiser, {\L}., Polosukhin, I.: Attention is all you need. In: Advances in
  neural information processing systems. pp. 5998--6008 (2017)

\bibitem{Warner:2012:DHS:2390374.2390377}
Warner, W., Hirschberg, J.: Detecting hate speech on the world wide web. In:
  Proceedings of the Second Workshop on Language in Social Media. pp. 19--26.
  LSM '12, Association for Computational Linguistics (2012)

\bibitem{wei2019eda}
Wei, J.W., Zou, K.: Eda: Easy data augmentation techniques for boosting
  performance on text classification tasks. arXiv preprint arXiv:1901.11196
  (2019)

\bibitem{wolf2019huggingface}
Wolf, T., Debut, L., Sanh, V., Chaumond, J., Delangue, C., Moi, A., Cistac, P.,
  Rault, T., Louf, R., Funtowicz, M., et~al.: Huggingface’s transformers:
  State-of-the-art natural language processing. ArXiv, abs/1910.03771  (2019)

\bibitem{wu2019conditional}
Wu, X., Lv, S., Zang, L., Han, J., Hu, S.: Conditional bert contextual
  augmentation. In: International Conference on Computational Science. pp.
  84--95. Springer (2019)

\bibitem{wu2016google}
Wu, Y., Schuster, M., Chen, Z., Le, Q.V., Norouzi, M., Macherey, W., Krikun,
  M., Cao, Y., Gao, Q., Macherey, K., et~al.: Google's neural machine
  translation system: Bridging the gap between human and machine translation.
  arXiv preprint arXiv:1609.08144  (2016)

\bibitem{xie2019unsupervised}
Xie, Q., Dai, Z., Hovy, E., Luong, M.T., Le, Q.V.: Unsupervised data
  augmentation. arXiv preprint arXiv:1904.12848  (2019)

\bibitem{yang2019xlnet}
Yang, Z., Dai, Z., Yang, Y., Carbonell, J., Salakhutdinov, R.R., Le, Q.V.:
  Xlnet: Generalized autoregressive pretraining for language understanding. In:
  Advances in neural information processing systems. pp. 5754--5764 (2019)

\bibitem{zhang2018word}
Zhang, D., Yang, Z.: Word embedding perturbation for sentence classification.
  arXiv preprint arXiv:1804.08166  (2018)

\bibitem{zhang2019paws}
Zhang, Y., Baldridge, J., He, L.: Paws: Paraphrase adversaries from word
  scrambling. arXiv preprint arXiv:1904.01130  (2019)

\end{thebibliography}
